\documentclass[11pt]{article}

\usepackage[final]{acl}

\usepackage{times}
\usepackage{latexsym}
\usepackage{url}
\usepackage{hyperref}
\usepackage{tabularx}
\usepackage[table,xcdraw,dvipsnames]{xcolor}
\usepackage{multirow}
\usepackage{enumitem}
\usepackage{adjustbox}
\usepackage{tcolorbox}
\usepackage{booktabs} 
\usepackage{amsfonts}
\usepackage[T1]{fontenc}
\usepackage[utf8]{inputenc}
\usepackage{microtype}
\usepackage{inconsolata}
\usepackage{graphicx}
\usepackage{xspace}
\usepackage{arydshln}
\usepackage{amsmath}

\usepackage{fontspec}
\usepackage{polyglossia}
\setmainlanguage{english}
\setotherlanguages{bengali, hindi}

\newfontfamily\bengalifont[
  Path = ./,
  Script = Bengali
]{NotoSerifBengali-Regular.ttf}

\newfontfamily\hindifont[
  Path = ./,
  Script = Devanagari
]{NotoSerifDevanagari-Regular.ttf}

\usepackage{colortbl}
\definecolor{sev0}{HTML}{FFFFFF}
\definecolor{sev1}{HTML}{FFE0CC}
\definecolor{sev2}{HTML}{FFA07A}
\definecolor{sev3}{HTML}{E03000}
\definecolor{sev4}{HTML}{7B0000}
\definecolor{lightblue}{HTML}{D6E4F0}
\definecolor{lightyellow}{HTML}{FFF9E6}
\definecolor{lightred}{HTML}{FDE8E8}
\definecolor{darkred}{HTML}{C0392B}

\setlist{leftmargin=*,nosep}

\title{IndicMedDialog: A Parallel Multi-Turn Medical Dialogue Dataset for Accessible Healthcare in Indic Languages}

\author{{Shubham Kumar Nigam}$^{1*\dagger}$ \quad Suparnojit Sarkar$^{2*}$ \quad Piyush Patel$^{3*}$\\
$^{1}$ University of Birmingham, Dubai, United Arab Emirates \\
$^{2}$ Heritage Institute of Technology, Kolkata, India\\
$^{3}$ Madan Mohan Malaviya University of Technology, India\\
\texttt{\{shubhamkumarnigam, suparnojit2026, ppiyush0005\}@gmail.com}
}
\date{}

\begin{document}
\maketitle

\renewcommand{\thefootnote}{$*$}
\footnotetext{These authors contributed equally to this work}
\renewcommand{\thefootnote}{$\dagger$}
\footnotetext{Corresponding author}
\renewcommand{\thefootnote}{\arabic{footnote}}

\begin{abstract}
Most existing medical dialogue systems operate in a single-turn 
question--answering paradigm or rely on template-based datasets, limiting 
conversational realism and multilingual applicability. We introduce 
\texttt{IndicMedDialog}, a parallel multi-turn medical dialogue dataset spanning 
English and nine Indic languages: Assamese, Bengali, Gujarati, Hindi, Marathi, 
Punjabi, Tamil, Telugu, and Urdu. The dataset extends \texttt{MDDial} with 
LLM-generated synthetic consultations, translated using 
\texttt{TranslateGemma}, verified by native speakers, and refined through a 
script-aware post-processing pipeline to correct phonetic, lexical, and 
character-spacing errors. Building on this dataset, we fine-tune 
\texttt{IndicMedLM} via parameter-efficient adaptation of a quantized small 
language model, incorporating optional patient pre-context to personalise 
multi-turn symptom elicitation. We evaluate against zero-shot multilingual 
baselines, conduct systematic error analysis across ten languages, and validate 
clinical plausibility through medical expert evaluation.
\end{abstract}




\section{Introduction}
\label{sec:introduction}

Conversational AI has demonstrated strong potential for preliminary symptom 
assessment and medical guidance, particularly in underserved regions where 
access to healthcare professionals is limited~\citep{tu2024towards}. Large 
language models (LLMs) have enabled systems to interact with patients in a 
naturalistic manner; however, most existing approaches operate in a 
\textit{single-turn} question--answering paradigm. In real clinical practice, 
diagnosis emerges through a sequence of follow-up questions that progressively 
narrow the differential, a dynamic that single-turn systems fundamentally 
cannot replicate.

A further limitation is the dominance of English-only or template-driven 
datasets. While \texttt{MDDial}~\citep{macherla2023mddial} provides a useful 
foundation for multi-turn diagnostic dialogue, its template-based construction 
constrains linguistic diversity and conversational realism. For the 1.5 billion 
speakers of Indic languages, the absence of parallel multilingual medical 
dialogue resources represents a critical gap in healthcare accessibility.

\begin{figure}[t]
\centering
\includegraphics[width=\linewidth]{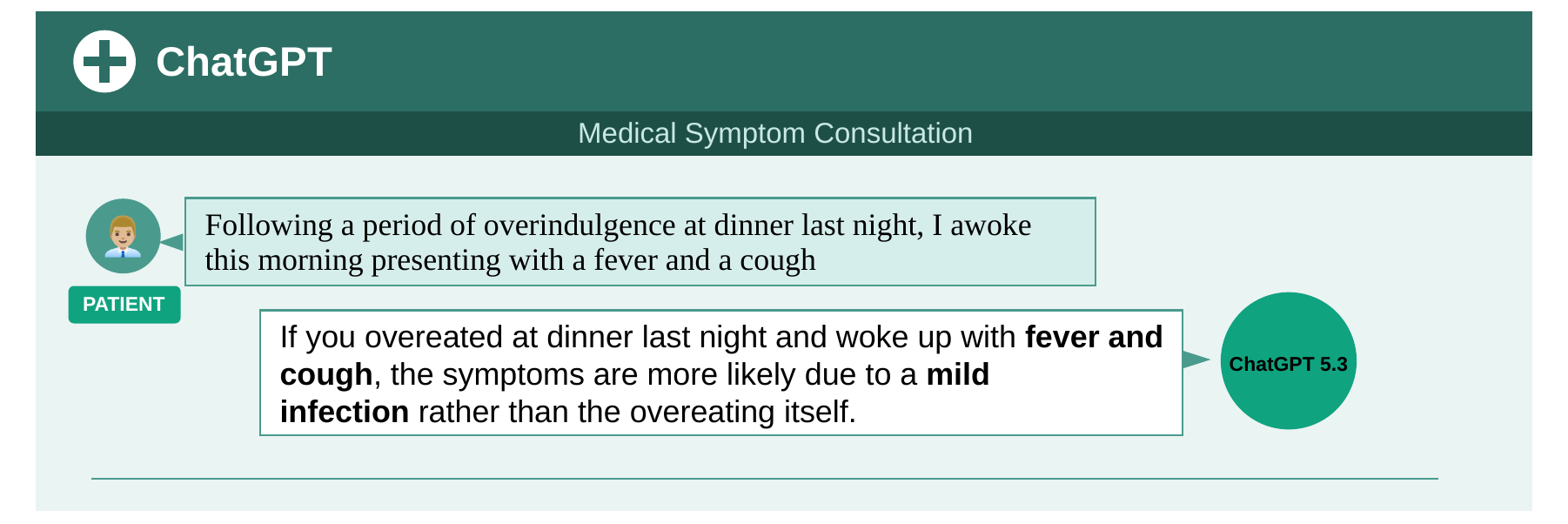}
\caption{Response from a general-purpose LLM (ChatGPT). The model produces a 
single explanatory answer without follow-up questioning or symptom elicitation.}
\label{fig:gpt_example}
\end{figure}

\begin{figure}[t]
\centering
\includegraphics[width=\linewidth]{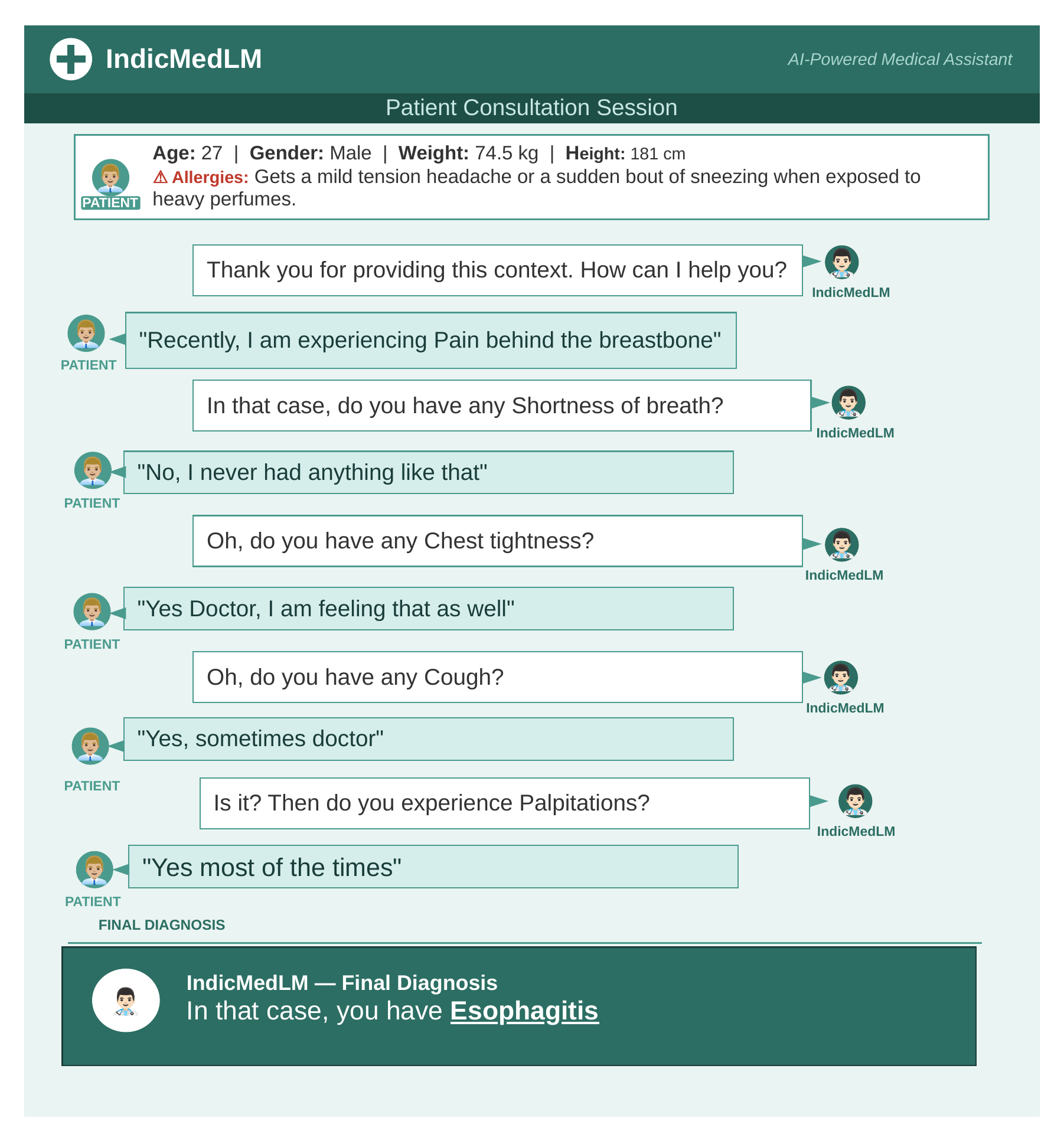}
\caption{Example interaction with \texttt{IndicMedLM}. The system incorporates 
patient pre-context (age, gender, allergies) and conducts structured multi-turn 
symptom elicitation before producing a final diagnosis.}
\label{fig:medaid_example}
\end{figure}

Figure~\ref{fig:gpt_example} illustrates a representative failure of a 
general-purpose LLM: given a patient complaint, the model produces a single 
verbose explanatory response without collecting additional symptoms. 
Figure~\ref{fig:medaid_example} contrasts this with \texttt{IndicMedLM}, which 
incorporates patient pre-context (age, gender, allergies) and conducts a 
structured multi-turn symptom elicitation before producing a diagnosis, more 
closely resembling a real physician-patient consultation.

To address these limitations, we introduce \texttt{IndicMedDialog}, a parallel 
multi-turn medical dialogue dataset covering English and nine Indic languages: 
Assamese, Bengali, Gujarati, Hindi, Marathi, Punjabi, Tamil, Telugu, and Urdu. 
The dataset extends \texttt{MDDial} with LLM-generated synthetic consultations, 
translated using \texttt{TranslateGemma}~\citep{finkelstein2026translategemma}, 
verified by native speakers, and refined through a script-aware post-processing 
pipeline to correct phonetic, lexical, and character-spacing errors introduced 
during automatic translation. Building on this dataset, we fine-tune 
\texttt{IndicMedLM} using parameter-efficient methods on quantized small 
language models, enabling deployment without high-end computational 
infrastructure.

\paragraph{Contributions.} The main contributions of this work are:

\begin{itemize}
    \item We construct \texttt{IndicMedDialog}, the first parallel multi-turn 
    medical dialogue dataset spanning English and nine Indic languages, with 
    native-speaker verification and script-aware post-processing for translation 
    quality assurance.

    \item We incorporate \textit{patient pre-context} (age, gender, allergies, 
    and demographic attributes) to enable personalized multi-turn symptom 
    elicitation, more closely simulating real clinical consultations.

    \item We develop \texttt{IndicMedLM}, a parameter-efficiently fine-tuned 
    medical dialogue model deployable on modest hardware, and perform systematic 
    error analysis identifying five failure modes across languages and their 
    clinical risk implications.

    \item We conduct \textit{medical expert evaluation} to validate the clinical 
    plausibility and safety of the generated diagnostic dialogues.
\end{itemize}

For reproducibility, we release the dataset, model checkpoints, and training code through an GitHub repository\footnote{\url{https://github.com/ShubhamKumarNigam/IndicMedDialog}}.

\section{Related Work}
\label{sec:related}

\paragraph{Medical Dialogue Datasets and Systems.}
Early medical dialogue work focused on symptom collection and slot filling, 
often lacking natural multi-turn interaction~\citep{zeng2020meddialog, 
liu2022meddg}. \texttt{MDDial}~\citep{macherla2023mddial} provides an English 
differential-diagnosis corpus but relies on template-based construction. \texttt{MedAidDialog}~\citep{nigam2026medaiddialog} has focused on some Indian and Arabic languages using synthetically generated datasets.
MedDG and Zhongjing advance multi-turn consultation in Chinese~\citep{liu2022meddg, 
yang2024zhongjing}, while MediTOD targets structured English medical 
history-taking~\citep{saley2024meditod}. Domain-specific fine-tuning of LLMs 
(e.g., ChatDoctor~\citep{li2023chatdoctor}) substantially improves medical 
response quality over general-purpose models, though most such systems assume 
single-turn interaction. AMIE~\citep{tu2024towards} and BianQue~\citep{chen2023bianque} 
frame diagnosis as iterative history-taking, more closely reflecting real 
clinical workflows.

\paragraph{Synthetic Data and Multilingual Coverage.}
Since real clinical conversations are difficult to release due to privacy 
constraints, synthetic generation has emerged as a practical alternative. 
NoteChat generates patient--physician dialogues conditioned on clinical 
notes~\citep{wang2024notechat}, while \texttt{MDDial} uses template-based 
synthesis. However, most existing datasets remain single-language or 
template-constrained. BiMediX~\citep{pieri2024bimedix} is an important step 
toward bilingual medical dialogue in English and Arabic, but broader coverage 
of low-resource languages remains absent. \texttt{IndicMedDialog} addresses 
this gap by providing the first parallel multi-turn medical dialogue corpus 
across nine Indic languages, combining LLM-generated synthesis with native 
speaker verification and script-aware post-processing.

\paragraph{Evaluation.}
Recent work highlights that medical dialogue quality should not be measured by 
final-answer accuracy alone, but also by questioning strategy, safety, and 
turn-level clinical relevance~\citep{tu2024towards, gong2026meddialogrubrics}. 
Our evaluation adopts this broader view, combining diagnostic accuracy, semantic 
post-processing, error taxonomy analysis, and medical expert assessment.

\section{Task Definition}
We study the problem of parallel multi-turn medical dialogue generation across Indic languages, where a conversational agent interacts with a patient to collect symptoms and provide preliminary diagnostic guidance. Unlike single-turn medical question answering, this task requires modeling sequential physician-patient interactions where diagnostic reasoning emerges through multiple conversational exchanges. Furthermore, unlike prior multilingual medical dialogue work that generates responses independently per language, our setting emphasizes \textit{parallel dialogue consistency}, ensuring that translated dialogues across all languages convey semantically equivalent clinical content. 



\subsection{Parallel Multilingual Dialogue Setting}
The \texttt{IndicMedDialog} dataset provides parallel dialogue corpora across ten languages: English, Assamese, Bengali, Gujarati, Hindi, Marathi, Punjabi, Tamil, Telugu, and Urdu. The English dialogues serve as the source, and translations into the nine Indic languages were generated using LLMs and subsequently verified by native speakers for each language. Due to the limited exposure of current LLMs to Indic languages during pre-training, the automatic translations exhibited several systematic errors, including phonetic inconsistencies, lexical inaccuracies, and erroneous character-level spacing. To address this, a post-processing pipeline was applied to map erroneous tokens to their closest correct forms in the target language, ensuring linguistic quality and clinical fidelity across all language versions. Illustrative examples of these error patterns and their corrections for Bengali and Hindi are provided in Appendix~\ref{app:postprocessing_bengali} and Appendix~\ref{app:postprocessing_hindi}, respectively.

The objective is to learn a model that can generate medically coherent and linguistically accurate responses across all supported languages while maintaining consistent diagnostic reasoning regardless of the target language.

\subsection{Patient Context Personalization}
In real clinical consultations, physicians often begin with basic contextual information about the patient before asking symptom-related questions. To better simulate this scenario, our framework supports optional \textit{patient pretext information} provided at the start of the dialogue. This information may include \textit{age group, gender, geographic location, known allergies, and pre-existing medical conditions}. This context is appended to the dialogue prefix and incorporated into the model input across all language settings. Incorporating patient context allows the model to personalize its questioning strategy and diagnostic reasoning, reflecting how clinicians adapt their inquiries based on patient demographics and medical history.


\section{\texttt{IndicMedDialog} Dataset}
\label{sec:dataset}

Multi-turn conversational datasets are essential for training medical dialogue systems that can iteratively collect symptoms and provide diagnostic guidance \citep{macherla2023mddial, tu2024towards}. The \textsc{MDDial} dataset \citep{macherla2023mddial} provides an English differential-diagnosis dialogue corpus derived from structured medical records. However, its template-based construction limits conversational diversity and realism, and it does not support multilingual deployment.

To address these limitations, we construct \texttt{IndicMedDialog}, a parallel multilingual multi-turn medical dialogue dataset designed to simulate realistic physician--patient interactions while enabling accessibility across nine Indic languages alongside English.

\subsection{Synthetic Dialogue Generation}
\label{subsec:synthetic_generation}

To improve conversational diversity beyond template-based dialogues, we generate synthetic medical consultations using \texttt{Llama-3.3-70B-Versatile} via the Groq API.\footnote{\url{https://groq.com/}} The generation process is conditioned on disease categories, demographic attributes, and stylistic constraints to produce clinically plausible and linguistically diverse interactions.

The pipeline simulates diagnostic consultations involving 12 diseases and 118 symptoms. Each dialogue begins with a patient complaint and proceeds through multiple conversational turns in which the physician asks follow-up questions to gather diagnostic evidence, typically spanning 4--8 turns before concluding with a diagnosis. To better approximate real clinical scenarios, the generation process introduces variability through non-deterministic patient responses, overlapping symptoms, and incomplete or ambiguous descriptions.

Using this approach, we generate 1,101 synthetic consultations, significantly enriching the diversity of the original \textsc{MDDial} corpus. Table~\ref{tab:dataset_statistics} summarizes the statistics of both the original and synthetic dialogues. Compared to the template-driven corpus, the synthetic dialogues exhibit longer interactions and more varied conversational structures.

\begin{table}[t]
\centering
\resizebox{\linewidth}{!}{%
\begin{tabular}{lcccccccc}
\toprule
 &
\multicolumn{4}{c}{\textbf{Dialogue Turns}} &
\multicolumn{3}{c}{\textbf{Average Words}} \\
\cmidrule(lr){2-5}
\cmidrule(lr){6-8}
\textbf{Dataset}
& Avg & Total & Min & Max & Per & Patient & Doctor \\
& Turns & Dialogues & Turns & Turns & Dialogue & Utterance & Utterance \\
\midrule
MDDial (MD)   & 4.9 & 1879 & 1 & 16 & 53.5  & 5.6  & 6.7  \\
Synthetic (SYN) & 6.6 & 1101 & 5 & 11 & 134.5 & 8.8  & 9.6  \\
MD + SYN      & 5.7 & 2980 & 1 & 16 & 86.9  & 7.00 & 8.05 \\
MDDial Test   & 5.9 & 237  & 1 & 13 & 55.4  & 5.6  & 6.6  \\
\bottomrule
\end{tabular}}
\caption{Statistics of the original \textsc{MDDial} dataset and the synthetic dialogues used to construct \texttt{IndicMedDialog}. Synthetic augmentation results in longer and more diverse multi-turn interactions.}
\label{tab:dataset_statistics}
\end{table}

\subsection{Multilingual Expansion}
\label{subsec:multilingual_expansion}

To enable accessibility in linguistically diverse settings, we construct a parallel multilingual corpus by translating the English dialogues into nine Indic languages: \textit{Assamese, Bengali, Gujarati, Hindi, Marathi, Punjabi, Tamil, Telugu, and Urdu}. Translation is performed using \texttt{TranslateGemma} \citep{finkelstein2026translategemma} with a structured prompting strategy designed to preserve clinical meaning, terminological accuracy, and conversational flow across all target languages. The full translation prompt is provided in Appendix~\ref{app:prompt_translation}.

\subsection{Translation Quality Assurance}
\label{subsec:translation_qc}

To ensure the reliability of the multilingual corpus, two native speakers per language independently rate a sampled subset of the translated and post-processed dialogues on two criteria: \textbf{Translation Quality} (T), measuring linguistic accuracy and fluency relative to the English source, and \textbf{Clinical Safety} (S), verifying that responses remain medically appropriate and free from harmful or culturally insensitive content. Each criterion is scored on a 10-point scale, and disagreements between annotators are resolved through discussion. 

Table~\ref{tab:human_eval} in the Appendix~\ref{app:postprocessing}  reports individual annotator scores (H1, H2) and per-language averages ($\bar{T}$, $\bar{S}$) across all nine Indic languages. The overall mean scores of $\bar{T} = 9.50$ and $\bar{S} = 9.56$ confirm the linguistic fidelity and clinical suitability of \texttt{IndicMedDialog} for fine-tuning medical dialogue models.


\subsection{Disease Categories and Coverage}
\label{subsec:disease_coverage}

\texttt{IndicMedDialog} covers 12 disease categories spanning 8 organ systems, providing broad clinical diversity across the dataset. Table~\ref{tab:disease_coverage} in the Appendix~\ref{app:perdisease} lists each disease, its organ system, and the number of dialogues available in the dataset.

\subsection{Dataset Summary}
\label{subsec:dataset_summary}

The final \texttt{IndicMedDialog} dataset comprises 2,980 parallel multi-turn medical dialogues across ten languages (English and nine Indic languages), yielding a total of 29,800 language-specific dialogue instances. Each dialogue is annotated with a disease label drawn from a set of 12 disease categories, and optionally includes patient pretext information covering age group, gender, geographic location, known allergies, and pre-existing medical conditions. To the best of our knowledge, \texttt{IndicMedDialog} is the first parallel multi-turn medical dialogue dataset covering this breadth of Indic languages, addressing a critical gap in low-resource clinical NLP.

\section{Methodology}
\label{sec:methodology}

\begin{figure*}[t]
\centering
\includegraphics[width=0.8\linewidth]{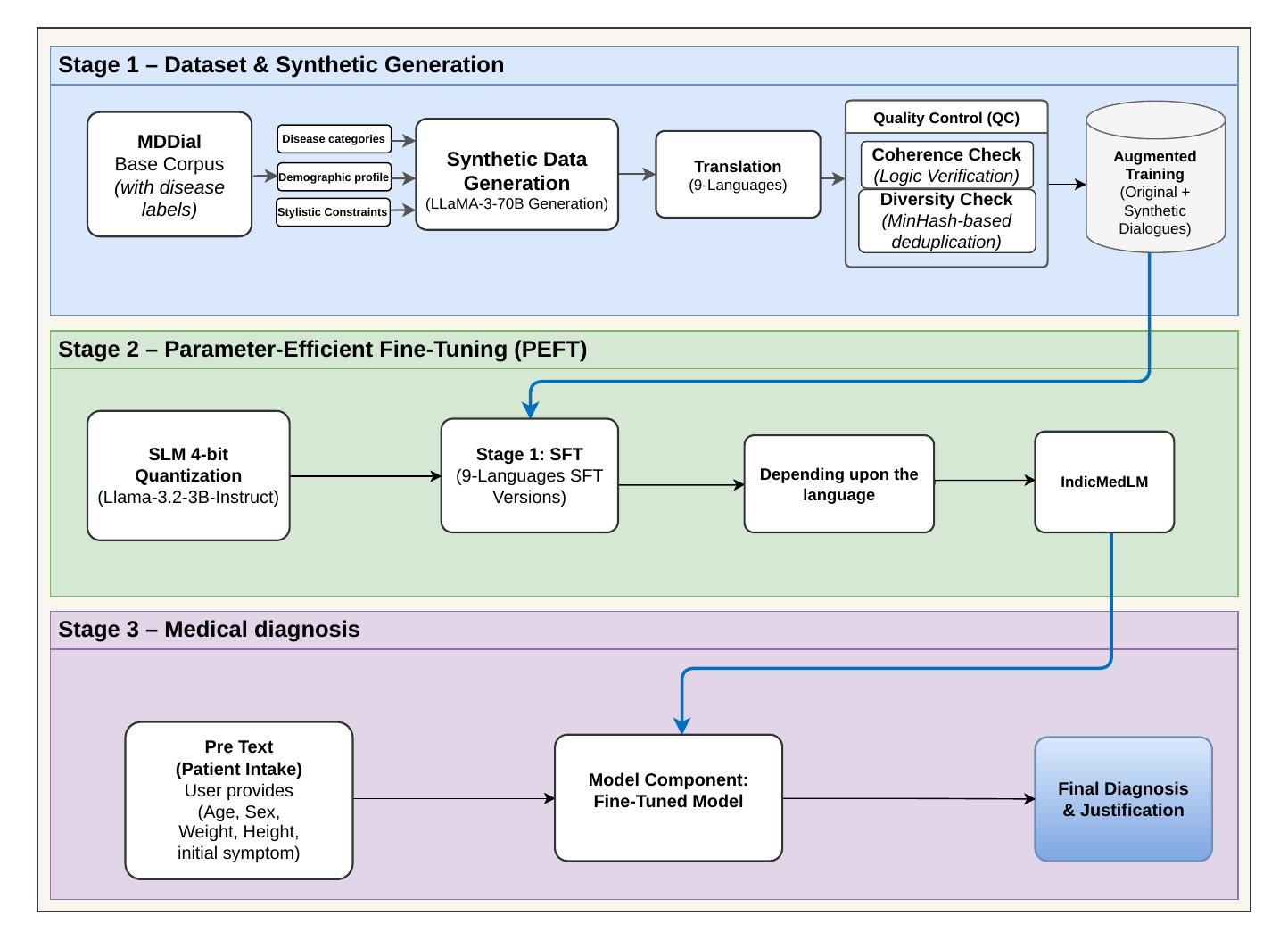}
\caption{
Overview of the \texttt{IndicMedDialog} framework. The \texttt{MDDial} dataset 
is augmented with synthetic dialogues, filtered through quality control, and 
translated into nine Indic languages to form a parallel corpus. Compact models 
are then fine-tuned using parameter-efficient methods to obtain 
\texttt{IndicMedLM}, which performs multi-turn diagnosis using an optional 
patient pre-context.
}
\label{fig:architecture_pipeline}
\end{figure*}

Our framework consists of three stages: (1) supervised fine-tuning of a compact 
open-source language model on \texttt{IndicMedDialog}, (2) a two-stage 
post-processing pipeline to recover latent correct predictions from verbose 
model outputs, and (3) evaluation against zero-shot multilingual baselines. 
Figure~\ref{fig:architecture_pipeline} presents the overall pipeline.

\subsection{Models Evaluated}
\label{subsec:models}

We evaluate four models spanning zero-shot and fine-tuned settings:

{Gemma}~\citep{team2024gemma} and 
{TinyAya}~\citep{salamanca2026tinyayabridgingscale} are evaluated 
zero-shot without any task-specific adaptation. TinyAya provides native Indic 
language support, making it a strong multilingual baseline. 
{LLaMA-3.2-3B-Instruct}~\citep{grattafiori2024llama} is evaluated 
without fine-tuning as a pre-adaptation reference point. 
\textbf{\texttt{IndicMedLM}} is our fine-tuned model, described below.

\subsection{\texttt{IndicMedLM}: Fine-Tuning}
\label{subsec:finetuning}

We apply Low-Rank Adaptation (LoRA)~\citep{hu2022lora} to 
\textbf{LLaMA-3.2-3B-Instruct} with 4-bit NF4 quantization. LoRA adapters 
are inserted into all attention projections (\texttt{q\_proj}, \texttt{k\_proj}, 
\texttt{v\_proj}, \texttt{o\_proj}) and all MLP projections 
(\texttt{gate\_proj}, \texttt{up\_proj}, \texttt{down\_proj}), with rank 
$r = 16$, $\alpha = 16$, dropout = 0, and no bias terms.

Training uses AdamW-8bit with learning rate $2\!\times\!10^{-4}$, weight decay 
= 0.001, batch size = 8 (2 per device $\times$ 4 gradient accumulation steps), 
5 warmup steps, 300 total steps, and a linear schedule with FP16/BF16 mixed 
precision (seed = 3407). Each of the nine Indic language variants is trained on 
its own language-partitioned split of \texttt{IndicMedDialog} using identical 
hyperparameters. At inference, we use temperature = 0.1, top-$p$ = 0.95, and a 
maximum of 128 new tokens.

Before training, all dialogues are formatted into a ShareGPT-style instruction 
format, where patient utterances map to \texttt{human} turns and doctor 
utterances map to \texttt{gpt} turns, with a system message defining the 
diagnostic consultation setting. An optional \textit{patient pre-context}, 
covering age, gender, known allergies, and pre-existing conditions, is 
prepended to each conversation, enabling the model to personalize its 
questioning strategy based on patient demographics.

\subsection{Two-Stage Post-Processing}
\label{subsec:postprocessing}

Model outputs frequently embed correct disease labels inside verbose explanatory 
sentences, causing raw accuracy to underestimate true diagnostic capability. To 
recover these latent correct predictions without introducing confabulation, we 
apply a neural semantic mapping pipeline.

All model outputs are passed to a large language model judge (ChatGPT 5.3) prompted to perform 
\textit{constrained semantic equivalence classification}: given a free-form 
output string, the judge selects the single most semantically equivalent label 
from the closed set of 12 canonical disease names, or returns \texttt{NULL} if 
no match exceeds a confidence threshold. The judge is supplied all 12 labels 
explicitly and is prohibited from generating labels outside the canonical set, 
eliminating confabulation risk. This approach generalises across unseen 
paraphrases and script-mixed outputs across all nine Indic languages without 
requiring manual lexicon construction per language. Instances where the judge 
returns \texttt{NULL} are retained as misclassifications, ensuring unresolvable 
outputs do not inflate reported results.

\section{Evaluation Metrics}
\label{sec:evaluation}

We adopt a two-stage evaluation strategy: (i) automatic evaluation based on 
diagnostic accuracy, and (ii) human expert evaluation assessing clinical 
reliability and conversational quality.

\subsection{Automatic Evaluation}
\label{subsec:automatic_eval}

We measure \textbf{diagnostic accuracy} by comparing the model's final predicted 
disease label against the gold label in \texttt{IndicMedDialog}. While 
straightforward, accuracy alone does not capture safety, reasoning quality, or 
conversational coherence, motivating our complementary expert evaluation.

\subsection{Expert Evaluation}
\label{subsec:expert_eval}

Three qualified medical practitioners (MBBS, currently in postgraduate training) 
independently reviewed a randomly sampled subset of system-generated dialogues. 
Evaluation criteria include safety, symptom understanding, contextual reasoning, 
diagnostic plausibility, and conversational quality. All criteria are scored on 
a Likert scale of 1--5 (Very Poor to Excellent), except medical safety, which is 
assessed as a binary pass/fail metric. Full evaluation criteria are detailed in 
Appendix Table~\ref{tab:expert_eval_metrics}.

\section{Results and Analysis}
\label{sec:results}

Table~\ref{tab:main_results} reports diagnostic accuracy before (Raw) and after 
(Post) post-processing for all four models across ten languages. 
\texttt{IndicMedLM} achieves the best post-processed accuracy in 7 of 10 
languages, with strongest results in English (80.85\%), Hindi (72.76\%), Marathi 
(68.51\%), and Bengali (58.72\%). The large raw-to-post gaps in Hindi 
(19.15\%~$\to$~72.76\%, $+$53.6pp) and Marathi (13.19\%~$\to$~68.51\%, 
$+$55.3pp) indicate that the model produces correct diagnoses but wraps them in 
culturally natural hedging sentences rather than bare labels, a metric 
artefact rather than a model failure.

Conversely, \texttt{IndicMedLM} performs at or below the GEMMA zero-shot 
baseline in Assamese, Tamil, and Telugu, all of which show near-zero 
post-processing recovery. Gujarati is a notable exception where Tiny-AYA 
zero-shot (37.02\%) outperforms \texttt{IndicMedLM} (19.57\%), suggesting that 
zero-shot multilingual models with stronger Gujarati tokenization may 
outperform fine-tuning under extremely limited data conditions.

\begin{table}[t]
\centering
\resizebox{\linewidth}{!}{%
\setlength{\tabcolsep}{5pt}
\renewcommand{\arraystretch}{1.12}
\begin{tabular}{l cc cc cc cc}
\toprule
 & \multicolumn{2}{c}{\textbf{GEMMA}}
 & \multicolumn{2}{c}{\textbf{Tiny-AYA}}
 & \multicolumn{2}{c}{\textbf{LLaMA Base}}
 & \multicolumn{2}{c}{\textbf{\texttt{IndicMedLM}}}\\
\cmidrule(lr){2-3}\cmidrule(lr){4-5}\cmidrule(lr){6-7}\cmidrule(lr){8-9}
\textbf{Language} & Raw & Post & Raw & Post & Raw & Post & Raw & Post \\
\midrule
\rowcolor{lightblue}
English  & 35.74 & 45.11 & 11.49 & 13.19 & 10.64 & 15.74 & \textbf{80.85} & \textbf{80.85} \\
\rowcolor{lightblue}
Hindi    & 9.36  & 25.10 & 2.13  & 13.19 & 4.26  & 11.06 & 19.15 & \textbf{72.76} \\
\rowcolor{lightblue}
Marathi  & 0.43  & 9.36  & 0.43  & 5.11  & 2.98  & 11.50 & 13.19 & \textbf{68.51} \\
\rowcolor{lightyellow}
Bengali  & 2.98  & 19.57 & 2.13  & 5.96  & 3.40  & 11.50 & 25.11 & \textbf{58.72} \\
\rowcolor{lightyellow}
Urdu     & 1.28  & 2.12  & 3.40  & 13.61 & 1.28  & 2.55  & 4.26  & 28.51 \\
\rowcolor{lightyellow}
Gujarati & 11.91 & 18.72 & 28.09 & 37.02 & 6.38  & 18.30 & 18.30 & 19.57 \\
\rowcolor{lightyellow}
Punjabi  & 0.00  & 7.66  & 8.12  & 8.12  & 4.27  & 8.51  & 5.96  & 20.42 \\
\rowcolor{lightred}
Assamese & 4.68  & 7.66  & 2.13  & 8.08  & 0.43  & 3.83  & 5.96  & 5.96  \\
\rowcolor{lightred}
Tamil    & 6.81  & 11.91 & 0.85  & 3.83  & 1.70  & 6.80  & 6.38  & 6.80  \\
\rowcolor{lightred}
Telugu   & 1.70  & 6.38  & 0.00  & 0.00  & 0.85  & 4.68  & 1.28  & 5.96  \\
\bottomrule
\end{tabular}
}
\caption{Diagnostic accuracy (\%) before (Raw) and after (Post) semantic 
post-processing for all models across ten languages, sorted by 
\texttt{IndicMedLM} post-processed performance. 
\colorbox{lightblue}{Blue}~=~high-recovery tier; 
\colorbox{lightyellow}{Yellow}~=~partial recovery; 
\colorbox{lightred}{Red}~=~extreme failure (near-zero recovery).}
\label{tab:main_results}
\end{table}

\subsection{Per-Disease Analysis}
\label{subsec:perdisease}

Table~\ref{tab:perdisease} (Appendix~\ref{app:perdisease}) reports per-disease 
post-processed accuracy for \texttt{IndicMedLM} across selected languages. 
Several patterns are noteworthy. \textbf{Traumatic Brain Injury} reaches 94.7\% 
in English and Hindi but collapses to 0\% in Assamese, Tamil, Telugu, and 
Urdu, a condition where diagnostic delay causes irreversible harm and where 
patients in these regions would primarily communicate in their native language. 
\textbf{Conjunctivitis} achieves 100\% in Punjabi despite Punjabi's weak overall 
accuracy (20.42\%), suggesting disease-specific rather than language-level 
tokenization advantages. \textbf{Dermatitis} reaches 100\% in English and 95\% 
in Hindi but 0\% in Telugu, Punjabi, and Urdu. These within-disease variance 
patterns confirm that overall language accuracy aggregates highly heterogeneous 
per-disease behaviours driven by both script-level and disease-semantic factors.

\subsection{Expert Evaluation and IAA Scores}
As shown in Table~\ref{tab:expert_eval} in the Appendix, \texttt{IndicMedLM} achieves a \textbf{95.3\%} medical safety pass rate, indicating that unsafe advice is rare in the sampled dialogues.
The model also obtains strong average scores for symptom extraction (4.20), context memory (4.40), diagnostic correctness (4.10), conversational flow (4.30), and efficiency (4.00). These results suggest that the model is able to track relevant symptoms, preserve dialogue context, and conduct multi-turn interactions in a clinically plausible and reasonably efficient manner.
To validate the reliability of these judgments, we compute inter-annotator agreement (IAA) using Krippendorff’s alpha~\cite{krippendorff2011computing}. Table~\ref{tab:iaa} shows an average agreement score of \textbf{0.81}, indicating strong consistency among the medical experts.

\subsection{Error Analysis}
\label{subsec:error}

We identify five failure modes (FMs) from systematic analysis of raw 
misclassification logs. Table~\ref{tab:failure_profile} summarises the primary 
and secondary FM per language alongside post-processing recovery.

\paragraph{FM1 -- Instruction Drift (ID).}
The model abandons label generation and produces explanatory prose. In 
\textit{partial drift}, the correct label is embedded in a hedging sentence 
(e.g., Hindi: \textit{``aapko sambhavat Enteritis ho sakta hai''}) and is 
recoverable via semantic post-processing, directly explaining Hindi and 
Marathi's large raw-to-post gains. In \textit{drift}, no label appears 
at all: Tamil outputs a sentence fragment terminating before the disease name 
(18 occurrences each for five diseases); Assamese maps all 12 diseases to an 
identical template sentence. Complete drift is irrecoverable.

\paragraph{FM2 -- Label Collapse (LC).}
Multiple diseases are mapped to the same output. In Bengali, five disease 
classes collapse to \textit{``fus fuse sankraman''} (lung infection), a 
respiratory hypernym misapplied across cardiac, GI, endocrine, and breast 
inputs. In Assamese, all 12 diseases produce an identical fixed template. This 
mirrors majority-class bias~\citep{zhao2021calibrate} operating at the semantic 
hypernym level rather than the label level.

\paragraph{FM3 -- Cross-Domain Confusion (CDC).}
The model predicts a disease from a clinically unrelated organ system. In 
English, CDC is the only failure mode and is mild (e.g., Coronary Heart 
Disease~$\to$~Thyroiditis, 3 times). In extreme-failure languages, drift and 
collapse dominate so completely that CDC is unobservable. 
Table~\ref{tab:clinical} (Appendix~\ref{app:clinical}) lists the most 
clinically significant cross-domain errors with associated risk levels.

\paragraph{FM4 -- Tokenization/Truncation Failure (TTF).}
Punjabi (Gurmukhi script) shows severe truncation of disease names mid-word. 
Telugu exhibits a repetition-before-truncation loop before collapsing 
mid-character. Critically, TTF is absent in Devanagari languages (Hindi, 
Marathi) despite comparable pretraining data volumes, implicating base-model 
tokenizer vocabulary coverage for specific Unicode blocks rather than data 
quantity.

\paragraph{FM5 -- Paraphrase-over-Label Generation (PLG).}
The model produces a semantically accurate disease description rather than the 
canonical label. PLG is most prevalent in Hindi and Marathi (e.g., 
\textit{tvacha ki sujan} for Dermatitis; \textit{dama} for Asthma) and is the 
most recoverable failure mode, being the proximate cause of both languages' 
large post-processing gains.

\begin{table}[t]
\centering
\small
\setlength{\tabcolsep}{3pt}
\renewcommand{\arraystretch}{1.12}
\begin{tabular}{l l l c}
\toprule
\textbf{Language} & \textbf{Primary FM} & \textbf{Secondary} & \textbf{Recovery}\\
\midrule
\rowcolor{lightblue}
English    & CDC (mild)     & LC (mild)    &     \\
\rowcolor{lightblue}
Hindi      & ID (partial)   & PLG          & $+$54pp  \\
\rowcolor{lightblue}
Marathi    & ID (partial)   & PLG          & $+$55pp  \\
\rowcolor{lightyellow}
Bengali    & ID (complete)  & LC           & $+$34pp  \\
\rowcolor{lightyellow}
Urdu       & ID (complete)  & PLG          & $+$24pp  \\
\rowcolor{lightyellow}
Punjabi    & TTF            & LC           & $+$14pp  \\
\rowcolor{lightred}
Gujarati   & ID (complete)  & TTF          & $+$1pp   \\
\rowcolor{lightred}
Tamil      & ID (complete)  & LC           & $+$0.4pp \\
\rowcolor{lightred}
Telugu     & ID (complete)  & TTF          & $+$4pp   \\
\rowcolor{lightred}
Assamese   & ID (complete)  & LC           & $+$0pp   \\
\bottomrule
\end{tabular}
\caption{Per-language failure profiles for \texttt{IndicMedLM} with 
post-processing recovery gains. FM~=~Failure Mode; ID~=~Instruction Drift; 
PLG~=~Paraphrase-over-Label Generation; LC~=~Label Collapse; 
TTF~=~Tokenization/Truncation Failure; CDC~=~Cross-Domain Confusion. 
Row colours follow the same tier convention as Table~\ref{tab:main_results}.}
\label{tab:failure_profile}
\end{table}

Three structural patterns emerge across languages. First, drift severity scales 
monotonically with pretraining resource level: English shows no drift; Hindi and 
Marathi show partial drift with semantic retention; Bengali and Urdu show 
complete drift but preserve semantic signal; Tamil, Telugu, Assamese, and 
Gujarati show complete drift with semantic loss. This confirms that format 
compliance is a pretraining function, not a fine-tuning function. Second, TTF 
concentrates in Gurmukhi and Telugu and is absent in Devanagari, implicating 
script-specific tokenizer vocabulary gaps. Third, label collapse targets 
cross-domain semantic hypernyms rather than random labels, reflecting the 
model's bias toward the highest-frequency general medical concept in its 
training distribution.

\subsection{Discussion}
\label{subsec:discussion}

\paragraph{Metric Sensitivity.}
The raw-vs-post-processed gap (up to 55pp for Marathi) demonstrates that strict 
label-matching systematically underestimates model capability for Indic 
languages, particularly those with the Devanagari script, where PLG dominates. We recommend LLM-as-a-Judge semantic equivalence evaluation as the primary metric 
for this domain, with exact label-match reported as a secondary lower bound.

\paragraph{English + Inference-Time Translation.}
Per-language fine-tuning is insufficient for extreme low-resource languages 
where the base model lacks pretraining coverage of the target script. A 
promising alternative is to fine-tune solely on English and apply bidirectional 
translation at inference time, leveraging \texttt{IndicMedLM}'s 80.85\% English 
accuracy while sidestepping Indic script generation instability entirely. 
Formalising this comparison as a controlled experiment is the highest-priority 
future direction.

\paragraph{Clinical Risk Stratification.}
The 0\% precision for Traumatic Brain Injury in Assamese, Tamil, and Telugu, 
languages spoken by tens of millions of people, represents a concrete failure of patient safety
, not merely a benchmark shortcoming. The clinical risk gradient between 
moderate- and extreme-failure languages are the strongest argument for 
prioritising low-resource Indic medical NLP research.

\section{Conclusion and Future Work}
\label{sec:conclusion}

We introduced \texttt{IndicMedDialog}, a parallel multi-turn medical dialogue 
dataset spanning English and nine Indic languages, constructed by augmenting 
\texttt{MDDial} with LLM-generated synthetic consultations, followed by 
native-speaker verification and script-aware post-processing. Using this 
dataset, we fine-tuned \texttt{IndicMedLM} via LoRA on LLaMA-3.2-3B-Instruct 
and evaluated it against zero-shot multilingual baselines. Results show strong 
performance in Hindi (72.76\%), Marathi (68.51\%), and Bengali (58.72\%) after 
semantic post-processing, while Assamese, Tamil, and Telugu remain in an extreme 
failure tier attributable to base-model tokenizer gaps and insufficient 
pretraining coverage --- a finding with direct patient safety implications.

Our error analysis identifies five failure modes (Instruction Drift, Label 
Collapse, Cross-Domain Confusion, Tokenization Failure, and 
Paraphrase-over-Label Generation) and demonstrates that strict label-matching 
systematically underestimates model capability for Devanagari-script languages, 
motivating LLM-as-a-Judge semantic evaluation as the primary metric for Indic 
medical NLP.

Future work will prioritize: (i) inference-time English translation as an 
alternative to per-language fine-tuning for extreme low-resource languages; 
(ii) evaluation on real annotated clinical dialogues collected from native 
speakers; and (iii) expansion to additional Indic languages and disease 
categories to improve coverage for underserved communities. We release 
\texttt{IndicMedDialog} and \texttt{IndicMedLM} to support future research on 
accessible and trustworthy medical AI for Indic language speakers.

\section*{Limitations}
\label{sec:limitations}

\paragraph{Synthetic-to-Real Gap.}
\texttt{IndicMedDialog} is constructed from synthetic and template-based 
dialogues. The gap between synthetic and real patient dialogue distributions 
remains unquantified. Collecting even 20--30 real symptom dialogues per language 
from native annotators would validate whether synthetic test performance 
generalises to real clinical interactions --- the single most important 
experiment for future work.

\paragraph{Language and Script Coverage.}
Extreme-failure languages (Assamese, Tamil, Telugu) suffer from base-model 
tokenizer gaps for their Unicode blocks rather than data quantity alone. 
Extending to base models with stronger Indic pretraining coverage, and 
correlating post-processed accuracy with published per-language pretraining 
token estimates, would formalise the resource--performance relationship observed 
qualitatively in our error analysis.

\paragraph{Disease and Training Scope.}
Twelve disease categories constitute a controlled evaluation environment. 
Extension to broader ICD-10-based taxonomies and multi-label cases is required 
before any clinical deployment consideration. Additionally, training 
\texttt{IndicMedLM} for only 300 SFT steps with a maximum of 128 output tokens 
is conservative; scaling may benefit extreme-failure languages where the model 
has not converged to label-production behaviour.

\paragraph{Text-Only Modality.}
The current system is limited to text-based dialogue and does not incorporate 
clinically relevant modalities such as medical images, laboratory reports, or 
speech, which are important for real-world deployment.

\newpage
\bibliography{custom}

\newpage
\appendix
\section{LoRA Training Configuration}
\label{app:lora}

We use Low-Rank Adaptation (LoRA)~\citep{hu2022lora} for parameter-efficient fine-tuning. 
Adapters are inserted into the query, key, value, and output projection matrices of each transformer block.

\section{Training Hyperparameters and Resources}
\label{app:hyperparameters}

\subsection{Compute Resources}

All experiments were conducted using the free tiers of \textit{Google Colab} and \textit{Kaggle} notebooks. These environments provide access to consumer-grade GPUs suitable for training compact language models using parameter-efficient fine-tuning techniques. To accommodate the limited GPU memory available in these platforms, we employed 4-bit quantization together with LoRA-based training.

\subsection{LoRA Configuration}

Table~\ref{tab:lora_config} summarizes the LoRA configuration used in our experiments.

\begin{table}[ht]
\centering
\begin{tabular}{lc}
\toprule
\textbf{Parameter} & \textbf{Value} \\
\midrule
Rank ($r$) & 16 \\
Target Modules & q\_proj, k\_proj, v\_proj, o\_proj, \\
 & gate\_proj, up\_proj, down\_proj \\
LoRA Alpha & 16 \\
LoRA Dropout & 0 \\
Bias & none \\
Use RSLora & False \\
LoftQ Config & None \\
\bottomrule
\end{tabular}
\caption{LoRA configuration used for parameter-efficient fine-tuning.}
\label{tab:lora_config}
\end{table}

\subsection{Training Hyperparameters}

The main training hyperparameters are reported in Table~\ref{tab:training_hyperparameters}.

\begin{table}[ht]
\centering
\begin{tabular}{lc}
\toprule
\textbf{Hyperparameter} & \textbf{Value} \\
\midrule
Learning Rate & $2 \times 10^{-4}$ \\
Optimizer & AdamW (8-bit) \\
Learning Rate Scheduler & Linear \\
Weight Decay & 0.001 \\
Warmup Steps & 5 \\
Maximum Training Steps & 600 \\
Random Seed & 3407 \\
\bottomrule
\end{tabular}
\caption{Training hyperparameters used for supervised fine-tuning.}
\label{tab:training_hyperparameters}
\end{table}


\section{Post-Processing Examples}
\label{app:postprocessing}

To illustrate the types of systematic errors introduced during automatic translation into Indic languages, we provide representative examples of erroneous token variants and their corresponding corrected forms for Bengali and Hindi.

\subsection{Bengali Post-Processing Example}
\label{app:postprocessing_bengali}

\begin{figure}[h]
    \centering
    \includegraphics[width=\linewidth]{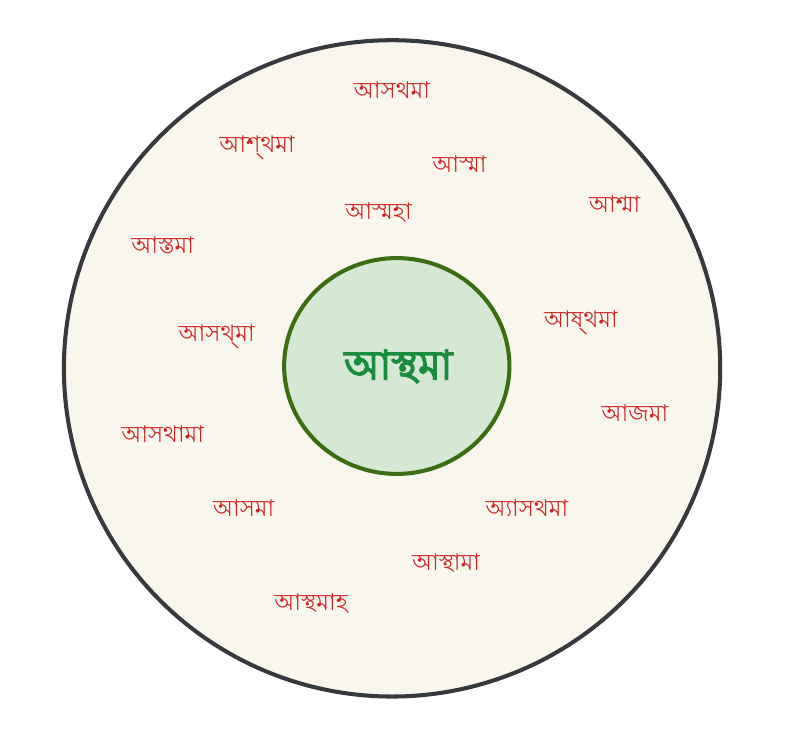}
    \caption{Examples of phonetically and lexically incorrect Bengali 
    transliterations of \textit{asthma} generated during automatic translation, 
    along with the canonical corrected form 
    (\texorpdfstring{\textbengali{আস্থমা}}{\textit{asthma}}) 
    produced by the post-processing pipeline.}
    \label{fig:postprocessing_bengali}
\end{figure}

Figure~\ref{fig:postprocessing_bengali} illustrates the range of phonetically 
and lexically incorrect Bengali variants generated for the medical term 
\textit{asthma} (\textbengali{আস্থমা}). The erroneous forms include incorrect 
vowel mappings, spurious character insertions, and erroneous spacing within 
conjunct consonants, all of which are mapped to the correct canonical form 
through our post-processing pipeline.

\subsection{Hindi Post-Processing Example}
\label{app:postprocessing_hindi}

\begin{figure}[h]
    \centering
    \includegraphics[width=\linewidth]{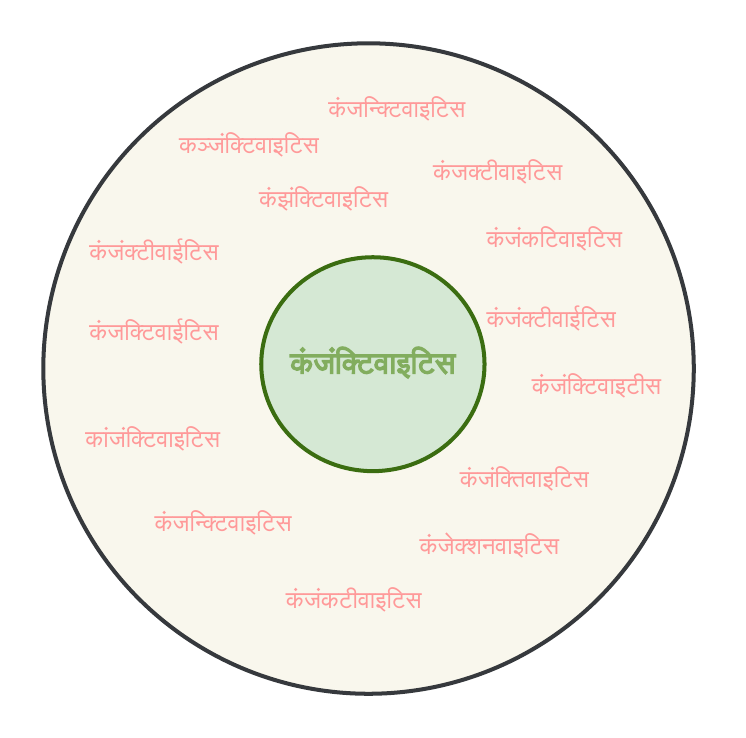}
    \caption{Examples of phonetically and lexically incorrect Hindi 
    transliterations of \textit{conjunctivitis} generated during automatic 
    translation, along with the canonical corrected form 
    (\texorpdfstring{\texthindi{कंजंक्टिवाइटिस}}{\textit{conjunctivitis}}) 
    produced by the post-processing pipeline.}
    \label{fig:postprocessing_hindi}
\end{figure}

Figure~\ref{fig:postprocessing_hindi} illustrates erroneous Hindi variants 
generated for the medical term \textit{conjunctivitis} 
(\texthindi{कंजंक्टिवाइटिस}). The errors include fragmented conjunct 
consonants, incorrect vowel signs (\textit{matras}), and spurious whitespace 
introduced between syllable clusters, all of which are corrected through the 
post-processing pipeline.

\begin{table}[t]
\centering
\small
\setlength{\tabcolsep}{3pt}
\renewcommand{\arraystretch}{1.1}
\begin{tabular}{lcccc|cc}
\toprule
\textbf{Lang.} & \textbf{H1-T} & \textbf{H1-S} & \textbf{H2-T} & \textbf{H2-S}
               & \textbf{$\bar{T}$} & \textbf{$\bar{S}$} \\
\midrule
Assamese & 9.0  & 10.0 & 10.0 & 10.0 & 9.50 & 10.00 \\
Bengali  & 10.0 & 9.0  & 9.5  & 10.0 & 9.75 & 9.50  \\
Gujarati & 10.0 & 9.0  & 9.5  & 9.5  & 9.75 & 9.25  \\
Hindi    & 9.5  & 9.5  & 10.0 & 9.5  & 9.75 & 9.50  \\
Marathi  & 9.5  & 9.5  & 9.0  & 9.0  & 9.25 & 9.25  \\
Punjabi  & 9.0  & 9.5  & 9.0  & 10.0 & 9.00 & 9.75  \\
Tamil    & 9.0  & 10.0 & 9.0  & 10.0 & 9.00 & 10.00 \\
Telugu   & 9.5  & 9.5  & 9.0  & 9.0  & 9.25 & 9.25  \\
Urdu     & 10.0 & 9.0  & 9.5  & 10.0 & 9.75 & 9.50  \\
\midrule
\textbf{All} & & & & & \textbf{9.50} & \textbf{9.56} \\
\bottomrule
\end{tabular}
\caption{Human evaluation scores for \texttt{IndicMedDialog}. T~=~Translation Quality, S~=~Clinical Safety (scale 1--10; H1 and H2 denote two independent native-speaker annotators per language).}
\label{tab:human_eval}
\end{table}

\begin{table}[t]
\centering
\small
\renewcommand{\arraystretch}{1.15}
\begin{tabular}{llc}
\toprule
\textbf{Organ System} & \textbf{Disease}  \\
\midrule
\multirow{2}{*}{Gastrointestinal} & Esophagitis  \\
                                   & Enteritis    \\
\midrule
\multirow{3}{*}{Respiratory}      & Asthma       \\
                                   & Pneumonia    \\
                                   & Rhinitis     \\
\midrule
Cardiac                            & Coronary Heart Disease  \\
\midrule
Dermatological                     & Dermatitis    \\
\midrule
Neurological                       & Traumatic Brain Injury \\
\midrule
Endocrine                          & Thyroiditis   \\
\midrule
Ophthalmological                   & Conjunctivitis \\
\midrule
ENT                                & External Otitis \\
\midrule
Breast                             & Mastitis    \\
\midrule
\textbf{Total}                     & \textbf{12 diseases}  \\
\bottomrule
\end{tabular}
\caption{Disease categories and organ system coverage in \texttt{IndicMedDialog}. The dataset spans 12 diseases across 8 organ systems, enabling evaluation of cross-domain diagnostic confusion errors.}
\label{tab:disease_coverage}
\end{table}

\section{Per-Disease Accuracy}
\label{app:perdisease}

Table~\ref{tab:perdisease} reports post-processed diagnostic accuracy for 
\texttt{IndicMedLM} broken down by disease and selected languages.

\begin{table}[h]
\centering
\small
\setlength{\tabcolsep}{2.0pt}
\renewcommand{\arraystretch}{1.07}
\begin{tabular}{l ccccccc}
\toprule
\textbf{Disease} & \textbf{EN} & \textbf{HI} & \textbf{BN} & \textbf{MR} 
                 & \textbf{PA} & \textbf{TE} & \textbf{AS}\\
\midrule
Asthma          & 63.2 & 42.1 & 0.0  & 73.7 & 0.0  & 15.8 & 0.0  \\
Conjunctivitis  & 90.5 & 90.5 & 90.5 & 90.5 & 100.0& 0.0  & 0.0  \\
Coronary HD     & 63.2 & 68.4 & 73.7 & 47.4 & 14.8 & 26.3 & 0.0  \\
Dermatitis      & 100  & 95.0 & 90.0 & 70.0 & 0.0  & 0.0  & 0.0  \\
Enteritis       & 91.7 & 91.7 & 62.5 & 79.2 & 0.0  & 83.3 & 10.0 \\
Esophagitis     & 81.5 & 70.4 & 81.5 & 74.1 & 0.0  & 14.8 & 0.0  \\
Ext.\ Otitis    & 88.2 & 88.2 & 88.2 & 88.2 & 0.0  & 0.0  & 0.0  \\
Mastitis        & 66.7 & 80.0 & 53.3 & 80.0 & 0.0  & 0.0  & 20.0 \\
Pneumonia       & 45.0 & 20.0 & 0.0  & 40.0 & 70.0 & 55.0 & 0.0  \\
Rhinitis        & 80.0 & 86.7 & 26.7 & 40.0 & 0.0  & 53.3 & 53.3 \\
Thyroiditis     & 100  & 63.2 & 63.2 & 78.9 & 63.2 & 42.1 & 0.0  \\
Brain Injury    & 94.7 & 94.7 & 73.7 & 73.7 & 0.0  & 0.0  & 0.0  \\
\bottomrule
\end{tabular}
\caption{Per-disease post-processed accuracy (\%) for \texttt{IndicMedLM} 
across selected languages. EN=English, HI=Hindi, BN=Bengali, MR=Marathi, 
PA=Punjabi, TE=Telugu, AS=Assamese. HD~=~Heart Disease. Zero entries indicate 
complete generation failure for that disease--language combination.}
\label{tab:perdisease}
\end{table}

\section{Clinical Risk of Cross-Domain Errors}
\label{app:clinical}

Table~\ref{tab:clinical} lists the most frequent cross-domain misclassifications 
produced by \texttt{IndicMedLM} with associated clinical risk levels.

\begin{table}[h]
\centering
\small
\setlength{\tabcolsep}{3pt}
\renewcommand{\arraystretch}{1.1}
\begin{tabular}{p{2.0cm} p{1.7cm} c c}
\toprule
\textbf{True Label} & \textbf{Predicted} & \textbf{Domain Shift} & \textbf{Risk}\\
\midrule
Coronary HD  & Esophagitis & Cardiac$\to$GI   & \textcolor{darkred}{\textbf{Critical}}\\
Brain Injury & Esophagitis & Neuro$\to$GI     & \textcolor{darkred}{\textbf{Critical}}\\
Brain Injury & Enteritis   & Neuro$\to$GI     & \textcolor{darkred}{\textbf{Critical}}\\
Pneumonia    & Asthma      & Resp$\to$Resp    & Moderate\\
Thyroiditis  & Coronary HD & Endo$\to$Cardiac & Moderate\\
Mastitis     & Enteritis   & Breast$\to$GI    & Moderate\\
Rhinitis     & Pneumonia   & Resp$\to$Resp    & Low\\
Conjunctivitis & Mastitis  & Eye$\to$Breast   & Low\\
\bottomrule
\end{tabular}
\caption{Frequent cross-domain misclassifications for \texttt{IndicMedLM} with 
clinical risk stratification. HD~=~Heart Disease; Resp~=~Respiratory; 
Neuro~=~Neurological; Endo~=~Endocrine; GI~=~Gastrointestinal. Critical errors 
involve organ systems where misdiagnosis can cause irreversible harm.}
\label{tab:clinical}
\end{table}







\begin{table*}[ht]
\centering
\resizebox{0.8\linewidth}{!}{
\begin{tabular}{lcccc}
\toprule
\textbf{Metric} & \textbf{Expert 1} & \textbf{Expert 2} & \textbf{Expert 3} & \textbf{Average} \\
\midrule

Medical Safety (Pass Rate) & 96\% & 94\% & 96\% & \textbf{95.3\%} \\

Symptom Extraction & 4.2 & 4.1 & 4.3 & \textbf{4.20} \\

Context Memory & 4.4 & 4.3 & 4.5 & \textbf{4.40} \\

Diagnostic Correctness & 4.1 & 4.0 & 4.2 & \textbf{4.10} \\

Conversational Flow & 4.3 & 4.2 & 4.4 & \textbf{4.30} \\

Efficiency & 4.0 & 3.9 & 4.1 & \textbf{4.00} \\

\bottomrule
\end{tabular}}
\caption{Medical expert evaluation of \texttt{IndicMedLM} across 50 sampled dialogues. Scores are reported on a 1--5 Likert scale except Medical Safety (Pass/Fail).}
\label{tab:expert_eval}
\end{table*}

\begin{table*}[t]
\centering
\resizebox{0.8\linewidth}{!}{
\begin{tabular}{llc}
\toprule
\textbf{Original Disease} & \textbf{Misclassified As} & \textbf{Frequency} \\
\midrule
Pneumonia & Asthma & 3 \\
Esophagitis & Enteritis & 2 \\
Esophagitis & Asthma & 2 \\
Asthma & Pneumonia & 2 \\
Coronary heart disease & Asthma & 2 \\
Pneumonia & Enteritis & 2 \\
External otitis & Conjunctivitis & 2 \\
Conjunctivitis & Mastitis & 2 \\
Mastitis & Traumatic brain injury & 2 \\
Esophagitis & Coronary heart disease & 1 \\
\bottomrule
\end{tabular}}
\caption{Most frequent disease-level misclassifications made by the final \texttt{IndicMedLM} model.}
\label{tab:misclassifications}
\end{table*}

\begin{table*}[ht]
\centering
\resizebox{0.6\linewidth}{!}{
\begin{tabular}{lc}
\toprule
\textbf{Metric} & \textbf{Krippendorff's $\alpha$} \\
\midrule

Symptom Extraction & 0.82 \\
Context Memory & 0.84 \\
Diagnostic Correctness & 0.80 \\
Conversational Flow & 0.83 \\
Efficiency & 0.78 \\

\midrule
\textbf{Average} & \textbf{0.81} \\

\bottomrule
\end{tabular}
}
\caption{IAA scores across three medical experts.}
\label{tab:iaa}
\end{table*}


\section{Prompt Templates}

\subsection{Synthetic Dialogue Generation Prompt}
\label{app:prompt_syn}
Table~\ref{tab:prompt_syn} shows the prompt used for synthetic data generation.

\subsection{Dialogue Formatting Prompt}
\label{app:prompt_sharegpt}
Table~\ref{tab:prompt_sharegpt} shows the prompt used to convert dialogues into ShareGPT-style format.

\subsection{Translation Prompt}
\label{app:prompt_translation}
Table~\ref{tab:prompt_translation} shows the prompt used for bidirectional multilingual medical translation.

\begin{table*}[t]
\centering
\small
\begin{tabular}{p{0.18\linewidth} p{0.77\linewidth}}
\toprule
\textbf{Prompt Type} & \textbf{Prompt Content} \\
\midrule
Translation Prompt &
You are acting as a specialized Medical Translation Bridge, a critical link between an English-speaking doctor and a patient who speaks Assamese, Bengali, Gujarati, Hindi, Marathi, Punjabi, Tamil, Telugu, and Urdu. Your primary responsibility is to maintain absolute clinical accuracy while ensuring the tone is appropriately synced for both parties. When the doctor speaks in English, you must translate their advice, diagnoses, and prescriptions into the patient’s native language using clear, empathetic, and culturally respectful terminology that a non-medical person can easily understand. Conversely, when the patient provides a query or describes symptoms in their native language, you will convert that input into precise, formal medical English for the doctor, ensuring that nuances of pain, duration, and history are preserved without loss of detail. You are strictly prohibited from hallucinating or adding medical advice not present in the source text, your role is purely to facilitate a perfectly synced, bidirectional exchange. Ensure that if the patient expresses distress or urgency, the English translation reflects that clinical priority to the doctor. Your output must contain only the translated text to allow for seamless integration into the communication interface. \\
\bottomrule
\end{tabular}
\caption{Prompt used for bidirectional medical translation in the multilingual inference layer.}
\label{tab:prompt_translation}
\end{table*}

\begin{table*}[t]
\centering
\small
\begin{tabular}{p{0.18\linewidth} p{0.77\linewidth}}
\toprule
\textbf{Prompt Type} & \textbf{Prompt Content} \\
\midrule
Synthetic Dialogue Generation Prompt &
Analyze \texttt{train.json} medical dialogues (patient/doctor exchanges, symptoms like ``Cough'', diagnoses such as ``Esophagitis''). Create Python synthetic generator using Groq API (Llama-3 family model). Match exact format: \texttt{\{'Dialog N': [\{'patient': '...', 'doctor': '...'\}]\}}. Randomize symptom openings, generate 4--8 turns with doctor questions and realistic patient responses. Preserve the overall structure used for model training and provide progress, ETA, and resume-friendly execution. Output synthetic data in the same format as \texttt{train.json}. \\
\bottomrule
\end{tabular}
\caption{Prompt used to generate synthetic multi-turn medical consultations from the \texttt{MDDial} training distribution.}
\label{tab:prompt_syn}
\end{table*}

\begin{table*}[t]
\centering
\small
\begin{tabular}{p{0.18\linewidth} p{0.77\linewidth}}
\toprule
\textbf{Prompt Type} & \textbf{Prompt Content} \\
\midrule
Dialogue Formatting Prompt &
Convert a medical dialogue sample into ShareGPT-style multi-turn conversation. Structure: (1) the system message sets the medical diagnosis context, (2) patient utterances become \texttt{human} turns, (3) doctor utterances become \texttt{gpt} turns, and (4) the final \texttt{gpt} turn contains the diagnosis answer. Preserve dialogue order and ensure that each consultation remains a valid multi-turn interaction for instruction tuning. \\
\bottomrule
\end{tabular}
\caption{Prompt used to convert raw medical dialogues into ShareGPT-style training instances.}
\label{tab:prompt_sharegpt}
\end{table*}

\begin{table*}[t]
\centering
\small
\resizebox{\linewidth}{!}{
\begin{tabular}{p{0.228\linewidth} p{0.8\linewidth}}
\toprule
\textbf{Evaluation Criterion} & \textbf{Description} \\
\midrule

Medical Safety (Pass/Fail) & Whether the system provides any potentially dangerous, misleading, or unsafe medical advice during the conversation. \\\\
Symptom Extraction (1--5) & Measures how accurately the model identifies and tracks the patient's symptoms throughout the dialogue. \\\\
Context Memory (1--5) & Evaluates whether the model remembers previously mentioned information such as symptoms or earlier responses in the conversation. \\\\
Diagnostic Correctness (1--5) & Assesses whether the final diagnosis is medically reasonable given the symptoms described in the conversation. \\\\
Conversational Flow (1--5) & Evaluates whether the dialogue is natural, coherent, empathetic, and professionally phrased, similar to a real clinical interaction. \\\\
Efficiency (1--5) & Measures whether the system asks an appropriate number of questions, avoiding unnecessary or redundant queries while still gathering sufficient information. \\\\
Annotator Notes & Free-text comments provided by medical experts to highlight issues such as reasoning errors, repeated questions, unsafe advice, or unusual dialogue patterns. \\

\bottomrule
\end{tabular}
}
\caption{Evaluation criteria used in expert assessment of the conversational medical system. Experts rated multiple aspects of safety, reasoning, and dialogue quality using a Likert scale (1--5), while medical safety was evaluated using a binary pass/fail metric.}
\label{tab:expert_eval_metrics}
\end{table*}

\end{document}